\documentclass{article}
\usepackage{algorithm}
\usepackage{algpseudocode}
\usepackage{amssymb}
\usepackage{bm}
\usepackage{algpseudocode}
\usepackage{enumitem}
\usepackage{booktabs}
\usepackage{makecell}
\usepackage{graphicx}
\usepackage{float}
\usepackage{wrapfig}
\usepackage{amsmath}
\usepackage{xcolor}
\usepackage{tabularx}
\usepackage{array}

\usepackage[final]{corl_2026} %
\usepackage{setspace}
\newcommand{\vth}{v_\theta}

\newcommand{\Vpsi}{V_\psi}
\newcommand{\hata}{\hat{a}}
\newcommand{\astar}{a^*}
\newcommand{\utstar}{u_t^*}
\newcommand{\EE}{\mathbb{E}}
\newcommand{\NN}{\mathcal{N}}
\newcommand{\DD}{\mathcal{D}}
\newcommand{\LL}{\mathcal{L}}
\newcommand{\RR}{\mathbb{R}}

\newcommand{\midrulefill}{%
  \leavevmode\leaders\hrule height0.8ex depth-0.65ex\hfill\kern0pt%
}

\title{FlowDPG: Deterministic Policy Gradient on Flow Matching Policies for Real-World Manipulation}

\author{
  \textbf{Kexin Shi}\textsuperscript{\normalfont 1,2} \quad
  \textbf{Junyao Shi}\textsuperscript{\normalfont 1,3} \quad
  \textbf{Poorvi Hebbar}\textsuperscript{\normalfont 1} \quad
  \textbf{Zhuolun Zhao}\textsuperscript{\normalfont 1} \\
  \vspace{1mm}
  \textbf{Tarun Amarnath}\textsuperscript{\normalfont 1} \quad
  \textbf{Yifan Su}\textsuperscript{\normalfont 1} \quad
  \textbf{Shikhar Bahl}\textsuperscript{\normalfont 1}\setcounter{footnote}{1}\thanks{Equal advising.} \quad
  \textbf{Deepak Pathak}\textsuperscript{\normalfont 1,2}\footnotemark[2] \\
  \vspace{2mm}
  \textsuperscript{1}Skild AI \quad
  \textsuperscript{2}Carnegie Mellon University \quad
  \textsuperscript{3}University of Pennsylvania
}

\begin{document}
\maketitle

\begin{figure}[H]
\centering
\includegraphics[width=\linewidth]{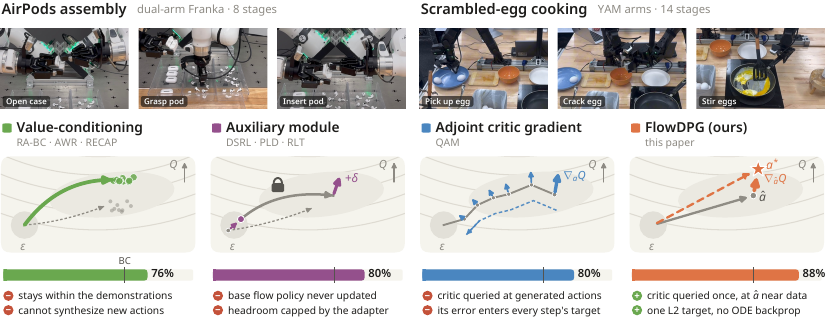}
\caption{\small \textbf{Top:} the two real-robot tasks. \textbf{Bottom:} 
how three families of RL fine-tuning methods (value-conditioning, 
auxiliary modules, and adjoint critic gradients) update a flow policy, 
their main limitations, and how FlowDPG avoids them. 
Bars: best offline AirPods success per family.}
\label{fig:teaser}
\end{figure}

\begin{abstract}
Real-world reinforcement learning for robotic manipulation remains 
challenging, and this difficulty is amplified for flow matching 
policies: policy gradients must be backpropagated through time (BPTT) 
along the multi-step ODE that maps noise to actions, which is 
computationally prohibitive and numerically fragile. We propose 
\textbf{FlowDPG}, a DDPG-style method for flow matching policies that 
bypasses BPTT entirely. FlowDPG evaluates the critic gradient at a 
one-step estimate of the clean action and distills the resulting 
correction into the velocity field at training time, leaving multi-step 
inference unchanged. Intuitively, it combines two complementary vectors: 
the demonstration-driven velocity that keeps the action feasible, and 
the critic-driven correction that steers it toward higher value. Our 
contributions are threefold: (1) a BPTT-free distillation framework for 
stable DDPG-style improvement of flow matching policies, (2) a formal 
connection between the FlowDPG update and the vanilla deterministic 
policy gradient via three explicit approximations, and (3) real-world 
validation on two long-horizon tasks on different robots. FlowDPG 
reaches 92\% end-to-end success on dual-arm AirPods assembly with Franka 
arms and an 86\% rubric score on scrambled-egg cooking with YAM arms, 
substantially outperforming recent RL methods. Videos and more results: 
\url{https://flowdpg.github.io}.
\end{abstract}


\raggedbottom

\setlength{\abovedisplayskip}{4pt plus 1pt minus 1pt}
\setlength{\belowdisplayskip}{4pt plus 1pt minus 1pt}
\setlength{\abovedisplayshortskip}{2pt plus 1pt}
\setlength{\belowdisplayshortskip}{2pt plus 1pt}

\setlength{\parskip}{2pt plus 0.3pt minus 0.3pt}

\setlength{\textfloatsep}{6pt plus 1pt minus 0pt}
\setlength{\floatsep}{6pt plus 1pt minus 0pt}
\setlength{\intextsep}{6pt plus 1pt minus 0pt}
\setlength{\abovecaptionskip}{2pt}
\setlength{\belowcaptionskip}{2pt}

\makeatletter
\renewcommand\section{%
  \@startsection{section}{1}{\z@}%
  {2ex \@plus 0.5ex \@minus 0.2ex}%
  {1ex \@plus 0.2ex}%
  {\normalfont\Large\bfseries}%
}
\renewcommand\subsection{%
  \@startsection{subsection}{2}{\z@}%
  {1.5ex \@plus 0.5ex \@minus 0.2ex}%
  {0.8ex \@plus 0.2ex}%
  {\normalfont\large\bfseries}%
}
\renewcommand\paragraph{%
  \@startsection{paragraph}{4}{\z@}%
  {1ex \@plus 0.3ex \@minus 0.1ex}%
  {-0.5em}%
  {\normalfont\normalsize\bfseries}%
}
\renewcommand{\ALG@beginalgorithmic}{\linespread{0.95}\selectfont}
\makeatother

\section{Introduction}
\label{sec:introduction}

Despite the successes of large-scale supervised models in robotics, imitation learning alone remains insufficient for long-horizon, contact-rich tasks. Collecting robot data at the scale and signal-to-noise ratio of the datasets used to train large language models is prohibitively expensive, and even with abundant high-quality teleoperation, a supervised policy inherits its demonstration distribution and degrades sharply in settings the data do not cover. On a long-horizon task, even rare failures at each stage compound into frequent end-to-end failure. This motivates letting the robot learn from its own experience. Real-world reinforcement learning, which has driven much of the recent progress in post-training large language models, is a natural candidate, as it can push the policy beyond the demonstration data.

Recent state-of-the-art manipulation policies 
have converged on complex generative approaches such as flow matching~\cite{lipman2023flowmatching} and 
diffusion~\cite{chi2023diffusion} action heads, which produce 
expressive multi-modal action chunks through a multi-step denoising 
ODE. For these policies, the standard off-policy actor-critic
update~\cite{lillicrap2016ddpg} propagating the critic gradient 
$\nabla_a Q$ back through the actor requires backpropagation 
through the entire denoising ODE. This is computationally 
prohibitive, since the cost grows linearly with the number of solver 
steps. It is also numerically fragile because the chain of 
Jacobians compounds gradient explosion and vanishing pathologies, 
particularly severe when the velocity field is not yet 
well-trained. As a result, expressive flow policies and 
value-driven RL have been hard to combine without either discarding 
the critic gradient or replacing the multi-step policy with a one-step 
actor.

We propose \textbf{FlowDPG}, an off-policy actor-critic method designed for flow 
matching policies that obtains a stable critic-gradient update 
without backpropagating through the denoising ODE. The key insight 
is that the clean action $x_1$ can be estimated from any 
intermediate noisy action $x_t$ by a single forward pass of the 
velocity predictor, as done for diffusion policies 
in EDP~\cite{kang2023edp}. We evaluate the critic gradient at this projected 
clean action, combine it with the demonstration-driven velocity to 
form a value-improved target, and distill the result back into the 
velocity field via L2 regression. Intuitively, FlowDPG composes 
two complementary vectors: the demonstration-driven velocity that 
keeps the action feasible, and the critic-driven correction that 
steers it toward higher value. Inference is unchanged from standard 
flow matching, a pure ODE solve with no deployment-time critic 
queries. A core property of FlowDPG is that its update direction can be 
derived from the vanilla DPG gradient on a flow policy under three 
explicit, controllable approximations, detailed in 
Section~\ref{sec:flowdpg}. This places the method on a transparent 
theoretical footing relative to the classical DDPG literature, in 
contrast to recent critic-gradient approaches that rely on more 
abstract stochastic-control formulations~\cite{li2026qam}.

We validate FlowDPG on two long-horizon real-world tasks on different 
robots. On contact-rich, dual-arm AirPods assembly with 
millimeter-level insertion tolerance, where a single policy runs all 
eight stages end-to-end from raw observations, FlowDPG reaches 92\% 
end-to-end success, 28 points above the BC base. On 14-stage 
scrambled-egg cooking with tool use and fluid handling, offline 
training alone raises the rubric score from 64.2\% to 85.8\%. It leads 
the closest baseline, QAM, on both tasks in a repeated evaluation, and 
online fine-tuning improves recovery from human disturbances.
In summary, our contributions are \textbf{(1)} a BPTT-free 
distillation framework for stable DDPG-style improvement of flow 
matching policies, \textbf{(2)} a formal connection between FlowDPG 
and vanilla DPG via three explicit approximations, and \textbf{(3)} 
real-world validation on two long-horizon tasks across two robot 
platforms, with ablations supporting each design choice.

\section{Related Work}
\label{sec:related_work}

\paragraph{Generalist manipulation policies and generative action heads.}
Large-scale robot datasets
\cite{o2023open,ebert2021bridge,robonet,khazatsky_droid_2024} have 
enabled generalist policies, from multi-task transformers
\cite{brohan2022rt1,brohan2023rt2,dasari2021transformers}
to vision-language-action models
\cite{kim2024openvla,black2024pi0,intelligence2025pi05,team2024octo,groot}
built on large vision-language models
\cite{touvron2023llama,chen2023pali,driess2023palme,karamcheti2024prismatic,beyer2024paligemma}.
At the action level, policies have moved to expressive generative 
heads: action-chunking transformers~\cite{act}, diffusion 
policies~\cite{chi2023diffusion,reuss2023goal,lbm}, and flow 
matching~\cite{lipman2023flowmatching,black2024pi0,groot}, the actor 
class we focus on.

\paragraph{RL fine-tuning of expressive manipulation policies.}
Recent work has explored several ways to improve imitation-trained
manipulation policies with reward signals. One family discards the
critic's action gradient and uses only scalar value or advantage
information: RA-BC~\cite{chen2025sarm} reweights the
behavior-cloning objective by progress estimates,
AWR~\cite{peng2019awr} fits the policy to advantage-weighted
demonstrations, and RECAP~\cite{intelligence2025recap} conditions the
policy on advantage as an additional input. A second family keeps the
base policy frozen and learns a separate adaptation module:
DSRL~\cite{wagenmaker2025dsrl} performs RL in the latent noise space
of the denoiser, PLD~\cite{xiao2025pld} learns a single-step residual
and distills it back through supervised fine-tuning, and
RLT~\cite{xu2026rlt} attaches a lightweight actor-critic head to a
compact token extracted from the frozen base policy. The closest
family exploits the critic's action gradient directly:
Q-score matching~\cite{psenka2024qsm} relates the score of an optimal
policy to gradients of the Q-function, while
QAM~\cite{li2026qam,wang2026lwd} converts $\nabla_a Q$ into
step-wise supervision through an auxiliary adjoint flow.
Diffusion-QL~\cite{wang2023diffusionql} backpropagates $Q$ through
the full denoising chain, and EDP~\cite{kang2023edp} replaces the
chain with a one-step estimate of the clean action. FlowDPG evaluates
the critic gradient at the analogous flow estimate, but normalizes the
step per sample, distills it as an L2 target, and derives the update
as an explicit approximation of vanilla DPG. On-policy methods avoid
BPTT by making each denoising step a Gaussian policy with a tractable
likelihood: DPPO~\cite{ren2025dppo}, ReinFlow~\cite{zhang2025reinflow},
and Flow-GRPO~\cite{liu2025flowgrpo} then apply PPO- or GRPO-style
updates, at the cost of on-policy data.
A parallel line avoids BPTT by changing the actor class: 
FQL~\cite{park2025fql} distills the flow into a one-step actor trained 
against $Q$, QC-FQL~\cite{li2025qchunking} extends it to action chunks, 
SSCQL~\cite{koirala2026sscp} and OFQL~\cite{nguyen2026ofql} train a 
single network for one-step generation, and the concurrent 
DriftQL~\cite{houssaini2026driftql} uses a one-step drift policy. All of them deploy a single-step generator. In the official FQL
setting, fresh noise is drawn at every control step and a single
action is executed, so closed-loop feedback absorbs the
stochasticity. When a one-step actor is executed in action chunks on
hardware, a fresh sample at each chunk boundary instead selects a
different trajectory, making chunk transitions jerky and
undeployable at millimeter tolerance, while fixing the noise removes
the multimodality the generative head was meant to provide. FlowDPG instead improves the multi-step flow policy in place. For value estimation, we use
IQL~\cite{kostrikov2021iql} with twin critics~\cite{fujimoto2018td3}
and a stage-aware reward predictor~\cite{chen2025sarm} as the dense
reward source.

\section{Method}
\label{sec:method}

\begin{figure*}[!t]
\centering
\includegraphics[width=\linewidth,
                 trim=183pt 208pt 183pt 208pt, clip]{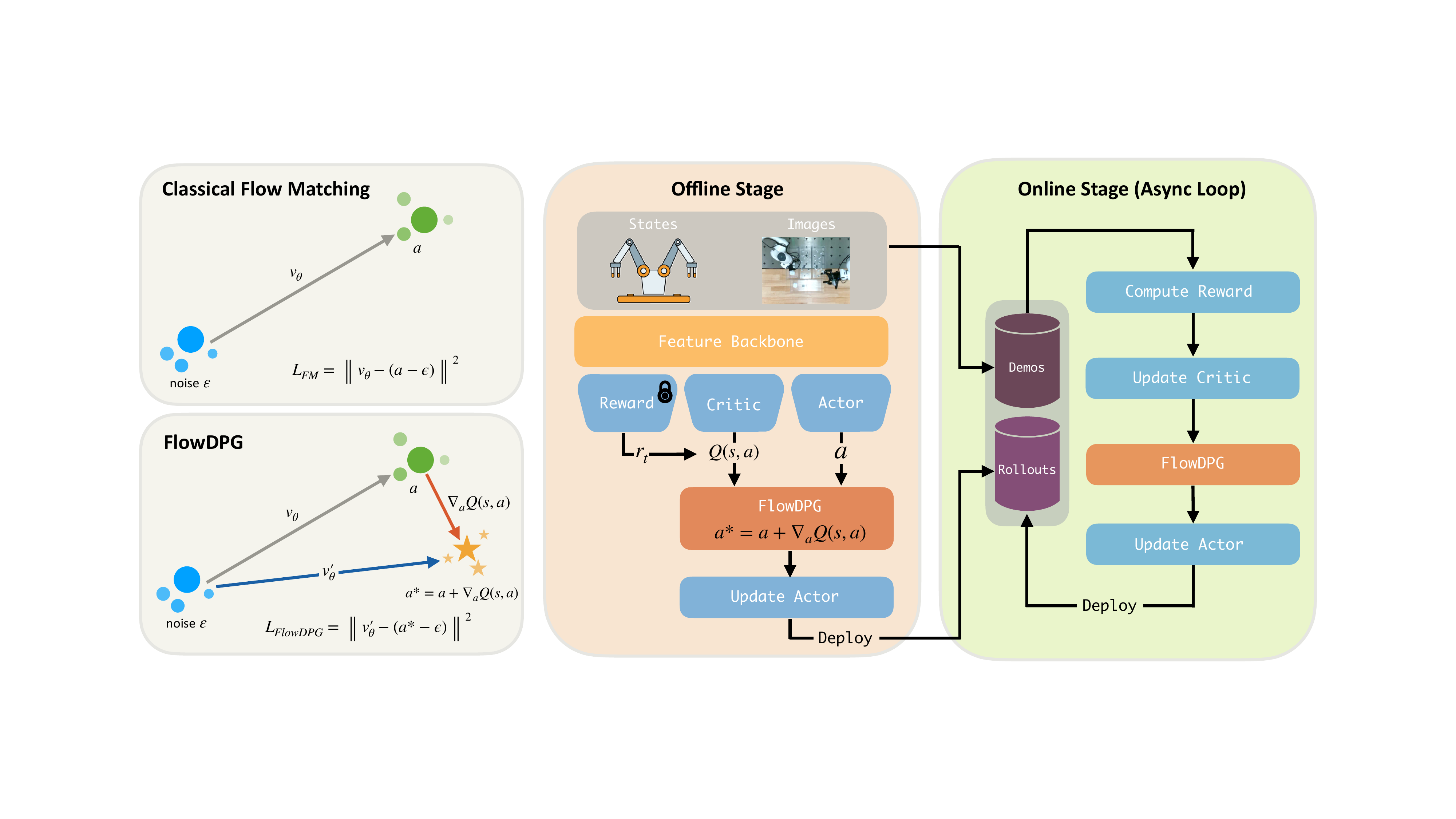}
\caption{\small \textbf{FlowDPG overview.} \textbf{Left:} classical 
flow matching (top) transports noise to a demonstration-feasible 
action $a$ via $\vth$; FlowDPG (bottom) adds a critic-driven 
correction along $\nabla_a Q$ to reach a value-improved target 
$\astar$ (schematic; Sec.~\ref{sec:flowdpg} queries the critic at the 
one-step estimate $\hata$ with an adaptive step), then distills the resulting velocity $\vth'$ back into the 
flow field. \textbf{Middle:} in the offline stage, a shared visual 
backbone feeds a frozen reward predictor, a critic, and a flow 
matching actor; the FlowDPG block combines the actor output with 
the critic gradient into $\astar$, which is distilled back into the 
actor. \textbf{Right:} in the online stage, 
the actor is rolled out on the real robot, and the critic and actor 
are asynchronously updated from a replay buffer mixing 
demonstrations with fresh rollouts.}
\label{fig:method}
\end{figure*}

Our method has two coupled components on top of a flow matching 
policy (Figure~\ref{fig:method}): a 
twin-critic IQL value estimator (Sec.~\ref{sec:value}), and 
\textbf{FlowDPG}, which distills the deterministic policy gradient 
(DPG)~\cite{silver2014dpg,lillicrap2016ddpg} into the flow matching 
velocity field at training time (Sec.~\ref{sec:flowdpg}).

\subsection{Value Estimation with Twin-Critic IQL}
\label{sec:value}

We adopt Implicit Q-Learning (IQL)~\cite{kostrikov2021iql} with twin 
critics in the style of TD3~\cite{fujimoto2018td3}. IQL replaces 
the $\max_a Q(s, a)$ bootstrap target, which queries the critic on 
potentially out-of-distribution actions, with a value function 
$\Vpsi(s)$ that estimates an upper-expectile in-distribution 
action-value at $s$, eliminating extrapolation error during value 
learning. Twin critics $Q_{\phi_1}, Q_{\phi_2}$ further mitigate 
Q-value overestimation from OOD actions and TD-bootstrap noise; we 
take their minimum wherever a single $Q$-estimate is needed, and 
pair each critic with an EMA target network used in the value loss. 
$\Vpsi$ is trained by expectile regression of the twin-critic 
minimum on dataset actions,
\begin{equation}
    \LL_V(\psi) = \EE_{\DD}\big[\rho_\tau\big(\min_{i=1,2} Q_{\bar\phi_i}(s, \bm{a}) - \Vpsi(s)\big)\big], 
    \quad \rho_\tau(u) = |\tau - \mathbb{1}[u<0]| \cdot u^2,
    \label{eq:vloss}
\end{equation}
with expectile $\tau > 0.5$ pulling $\Vpsi(s)$ toward the upper 
tail of the action-value distribution. The critics are then 
bootstrapped from $\Vpsi$:
\begin{equation}
    y_Q = r_t + \gamma^H \Vpsi(s_{t+H}), \qquad
    \LL_Q(\phi_i) = \EE_{\DD}\big[\big(Q_{\phi_i}(s_t, \bm{a}_t) - y_Q\big)^2\big],
    \label{eq:qloss}
\end{equation}
again avoiding extrapolation to unseen actions. Chunk-level rewards 
$r_t$ come from SARM~\cite{chen2025sarm}, a frozen stage-aware 
reward predictor that outputs a stage label $z(o_t)$ and a 
within-episode progress $\phi(o_t) \in [0, 1]$, combined into a 
composite chunk reward of three terms:
\begin{equation}
    r_t = \underbrace{\phi(o_{t+H}) - \phi(o_t)}_{\text{progress}} 
        \;+\; \underbrace{\beta_{\text{stage}} \cdot \mathbb{1}\big[z(o_{t+H}) > z(o_t)\big]}_{\text{stage\_transition}}
        \;+\; \underbrace{r_t^{\text{term}}}_{\text{terminal}}.
    \label{eq:chunkreward}
\end{equation}
The progress term provides dense per-step shaping, the indicator 
term issues a fixed bonus $\beta_{\text{stage}}$ at each predicted 
stage advance, and the terminal term encodes whole-episode 
success; we visualize an example reward and value curve in 
Appendix~\ref{app:sarm_visualization}.

\subsection{Policy Extraction via FlowDPG}
\label{sec:flowdpg}

A flow matching policy transports noise $x_0$ to a feasible action 
$x_1$ within the demonstration support, but the resulting action 
is not necessarily optimal under the critic. FlowDPG combines 
\emph{two vectors} (Figure~\ref{fig:method}, left): the 
demonstration-driven velocity for feasibility, and a critic-driven 
correction along $\nabla_a Q$ in the spirit of 
DDPG~\cite{lillicrap2016ddpg,silver2014dpg} for value improvement. 
The combined target is distilled back into $\vth$ via a single L2 
regression, avoiding the obvious alternative of backpropagating 
$\nabla_a Q$ through the full ODE, which is computationally prohibitive 
and numerically fragile. FlowDPG costs only one extra forward pass 
and one critic-gradient evaluation per update, and leaves 
inference as a pure flow ODE solve.

\paragraph{Q-Gradient.} The clean 
action $x_1$ can be estimated from any intermediate $x_t$ by a 
single forward pass of $\vth$:
\begin{equation}
    \hata := x_t + (1 - t) \cdot \vth(x_t, t, s).
    \label{eq:project}
\end{equation}
For an accurate $\vth$, $\hata$ equals $\EE[x_1 \mid x_t]$ and 
approaches $x_1$ as $t \to 1$; the consistency regularizer below 
reduces its fitting error. We compute $\hata$ under 
\texttt{no\_grad} and evaluate the critic gradient there:
\begin{equation}
    g = \nabla_{\hata}\!\left[\min\nolimits_{i=1,2} Q_{\phi_i}(s, \hata)\right],
    \label{eq:qgrad}
\end{equation}
with stop-grad on the critic parameters.

\paragraph{Adaptive shift.} The magnitude of $g$ varies 
considerably across states and training steps, destabilizing 
updates if used as-is. We normalize the step to the local velocity 
scale:
\begin{equation}
    \Delta = \alpha \cdot \frac{\|u_t\|}{\|g\| + \varepsilon} \cdot g,
    \label{eq:adaptive}
\end{equation}
with $\alpha > 0$ a base guidance scale and $\varepsilon$ guarding 
against vanishing gradients.

\paragraph{Distilled flow target.} Summing $\hata$ and $\Delta$ 
gives the Q-improved clean action, and subtracting noise gives the 
velocity target:
\begin{equation}
    \astar = \hata + \Delta, 
    \qquad \utstar = \astar - x_0.
    \label{eq:target}
\end{equation}
The policy is regressed against $\utstar$ alongside a 
behavior-cloning anchor:
\begin{equation}
    \LL_{\text{FlowDPG}}(\theta) = \lambda \cdot \underbrace{\|\vth(x_t, t, s) - \utstar\|^2}_{\LL_{\text{distill}}} + (1 - \lambda) \cdot \underbrace{\|\vth(x_t, t, s) - u_t\|^2}_{\LL_{\text{BC}}},
    \label{eq:flowdpgloss}
\end{equation}
with $\lambda$ warming up linearly from 0 over $N_{\text{warmup}}$ 
steps so that the unreliable early critic does not dominate.

\paragraph{Consistency regularizer.} The reliability of $\hata$ is 
critical: a projection off the clean-action manifold leaves the 
critic queried OOD and $g$ unreliable. Since the standard flow loss 
only supervises local velocity, we add, with weight $\mu_{\text{cons}}$,
\begin{equation}
    \LL_{\text{cons}}(\theta) = \EE\big[\, \| \hata - x_1 \|^2 \,\big].
    \label{eq:cons}
\end{equation}
Substituting $x_t = (1-t)x_0 + t x_1$ into Eq.~\ref{eq:project} 
gives $\hata - x_1 = (1-t)(\vth - u_t)$, so $\LL_{\text{cons}}$ 
reduces to a velocity-error penalty weighted by $(1-t)^2$, 
emphasizing small-$t$ samples where the projection error is most 
amplified.

\paragraph{Relationship to vanilla DPG.} A vanilla 
DPG~\cite{silver2014dpg,lillicrap2016ddpg} update on a flow matching 
policy treats the ODE terminus $x_1$ as a deterministic actor and, by 
the chain rule (Appendix~\ref{app:vanilla_dpg}), yields
\begin{equation}
    \nabla_\theta \LL_{\text{DPG}} 
    = -\nabla_a Q(s, x_1) \cdot \nabla_\theta x_1
    = -\int_0^1 g^\top \Phi(1, t) \cdot \nabla_\theta \vth(x_t, t, s)\, dt,
    \label{eq:vanilla_dpg}
\end{equation}
where $\Phi(1, t) := \partial x_1 / \partial x_t$ is the ODE 
sensitivity matrix. Computing $\Phi$ 
requires backpropagating through all $T$ ODE steps with $O(T)$ memory 
and well-known gradient explosion/vanishing pathologies, particularly 
severe when $\vth$ is not yet well-trained. The FlowDPG gradient, by 
contrast, comes from differentiating $\LL_{\text{FlowDPG}}$: after 
warmup, the only $g$-dependent term is the distillation loss; 
treating $g$ as a detached constant, its part of the gradient is
\begin{equation}
    \big[\nabla_\theta \LL_{\text{FlowDPG}}\big]_g 
    = -2\lambda \alpha \|u_t\| \cdot \hat{g}(\hata) \cdot \nabla_\theta \vth(x_t, t, s),
    \label{eq:flowdpg_grad_recap}
\end{equation}
with $\hat{g}(\hata) := g / \|g\|$. This recovers 
Eq.~\ref{eq:vanilla_dpg} under three explicit approximations: 
\emph{(i)} the critic gradient is evaluated at the one-step estimate 
$\hata$ rather than at the ODE terminus, collapsing $\Phi(1, t)$ to 
the identity and eliminating ODE backpropagation, with 
$\LL_{\text{cons}}$ reducing the error of this estimate; \emph{(ii)} the trajectory-wide 
integral is replaced by a Monte Carlo estimate at a single 
$t \sim \mathrm{Uniform}(0, 1)$, the same estimator flow matching uses 
for training; and \emph{(iii)} the un-normalized step size 
$\|g\|$ is replaced by the adaptive scaling 
$2\lambda\alpha\|u_t\|$ of Eq.~\ref{eq:adaptive}, removing a known 
source of instability in DDPG-family 
methods~\cite{lillicrap2016ddpg,fujimoto2018td3}.

\subsection{Offline-to-Online Pipeline}
\label{sec:pipeline}

The two components above slot into the pipeline of 
Figure~\ref{fig:method} (middle and right), detailed in Algorithm~\ref{alg:main} 
(Appendix~\ref{app:algorithm}). 
In the \textbf{offline phase}, the backbone, action head, and value 
head are jointly trained on demonstrations $\DD_{\text{off}}$ with 
the FlowDPG objective, producing a base policy that initializes 
online deployment. In the \textbf{online phase}, the policy is 
deployed to collect real-world rollouts that are asynchronously 
appended to $\DD_{\text{on}}$; the learner samples mini-batches at 
$\sim$1:1 ratio from $\DD_{\text{off}}$ and all rollouts in 
$\DD_{\text{on}}$ to balance demonstration quality and fresh 
experience. The BC anchor $\LL_{\text{BC}}$ is computed only on 
demonstrations and \emph{successful} rollouts, so it stays tied to 
successful behavior rather than drifting with the policy. Backbone and 
reward predictor are frozen in this phase to preserve pretrained 
representations and reduce per-step compute; updated actor/critic 
checkpoints synchronize back to deployment periodically.

\section{Experiments}
\label{sec:expe}

\begin{figure*}[t]
\centering
\includegraphics[width=\linewidth]{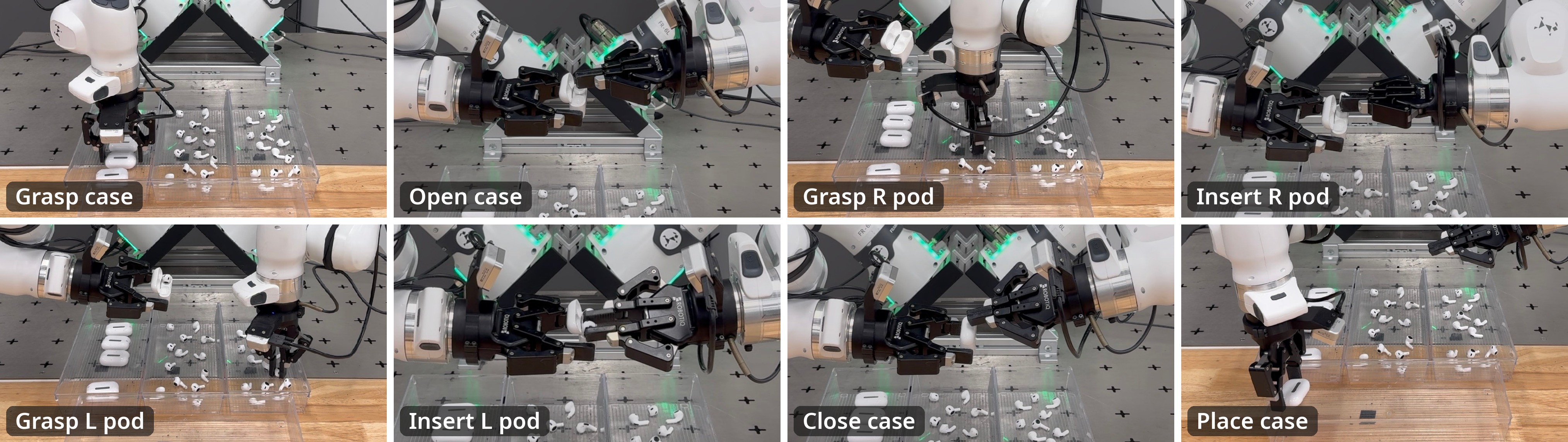}\\[1pt]
{\small (a) Dual-arm AirPods assembly (Franka), 8 stages}\\[5pt]
\includegraphics[width=\linewidth]{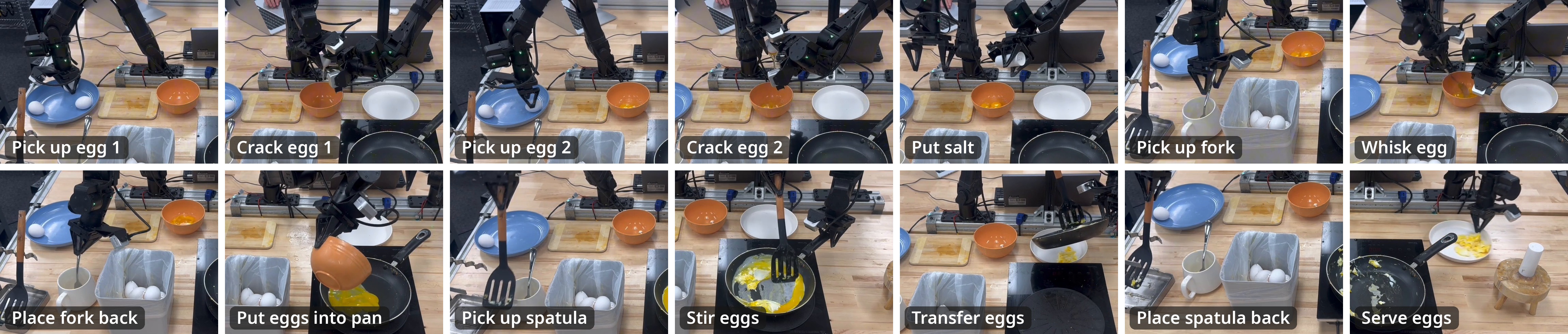}\\[1pt]
{\small (b) Scrambled-egg cooking (YAM), 14 stages}
\caption{\small The two long-horizon tasks, stage by stage (left to 
right, top to bottom; egg keyframes from one autonomous rollout). A 
single policy executes all stages end-to-end from raw observations, 
without stage-level resets or hand-engineered switching.}
\label{fig:tasks}
\end{figure*}

\paragraph{Tasks.} We evaluate on two long-horizon tasks on different 
robots (Figure~\ref{fig:tasks}); demonstrations for both are collected 
with GELLO~\cite{wu2024gello} teleoperation. \emph{AirPods assembly} 
runs on two Franka arms with randomized case and pod poses and has 8 
stages: grasp the case, open it, grasp and insert the right pod, grasp 
and insert the left pod, close the case, and place it. The policy sees 
four cameras (front, top, and two wrist views). \emph{Scrambled-egg 
cooking} runs on two YAM arms and has 14 stages: crack two eggs, add 
salt, fetch a fork to whisk and return it, pour into a pan, fetch a 
spatula to stir and transfer the eggs, return it, and serve. It adds 
tool use, fluid handling, and exogenous dynamics, and the policy sees 
three cameras (top and two wrist views).

\paragraph{Metrics and protocols.} The \emph{full sweep} 
(Table~\ref{tab:main_airpods}) evaluates every method in one round of 
25 AirPods cases (50 pod manipulations); failures within a stage do 
not end the episode, retries count in the per-stage denominators, and 
\textbf{Overall} requires all pods correctly inserted. The 
\emph{repeated evaluation} (Figure~\ref{fig:main_results}) compares BC, 
QAM, and FlowDPG on both tasks with 50 rollouts per cell, 10 at each of 
5 checkpoints of one training run ($\pm$: s.e.\ over checkpoint means), 
and adds a \textbf{rubric} score grading each stage 0--3 as in 
$\pi_{0.5}$~\cite{intelligence2025pi05} (Appendix~\ref{app:rubric}). 
The egg task is evaluated after the offline phase only.

\paragraph{Baselines.} We compare against three families of RL
methods for adapting a base flow policy. \emph{Value-conditioning}
methods use only scalar value/advantage signals:
RA-BC~\cite{chen2025sarm} reweights the BC objective by SARM
progress, AWR~\cite{peng2019awr} performs
advantage-exponentially-weighted regression, and
RECAP~\cite{intelligence2025recap} conditions the policy on
advantage. \emph{Auxiliary-module} methods keep the base policy
frozen and train a separate adapter:
DSRL~\cite{wagenmaker2025dsrl} performs RL in the noise space of
the denoiser, PLD~\cite{xiao2025pld} learns a single-step residual,
and RLT~\cite{xu2026rlt} attaches an actor-critic head to a
compact RL token. The \emph{critic-gradient} category contains the
concurrent QAM~\cite{li2026qam}, which converts $\nabla_a Q$ into
step-wise supervision via adjoint matching. All baselines share the
same backbone, action chunk horizon, reward, value head, and
offline data as FlowDPG. 

\subsection{Main Results}
\label{sec:q1}

\paragraph{Value-conditioning is bounded by the demonstration 
support.} Reweighting or conditioning over demonstrations cannot 
synthesize an action outside their distribution: these methods help 
where the bottleneck is picking the better mode (Open/Close case) but 
stay near BC on contact-rich stages (Insert R/L pod).

\paragraph{Auxiliary modules leave the base policy untouched.} DSRL 
acts in noise space, separated from the action by the entire ODE; 
PLD's single-step residual nudges the endpoint but cannot reshape the 
denoising; RLT is strongest here, but its MLP actor discards the flow 
policy's multimodality. Headroom is capped by the adapter.

\paragraph{Critic-gradient methods update the base policy.} FlowDPG 
uses $\nabla_a Q$ to update the velocity field itself, removing both 
ceilings. The concurrent QAM~\cite{li2026qam} pursues the same 
direction through adjoint matching but supervises every flow 
timestep, so critic error at the generated action enters every step's 
target; FlowDPG distills a single critic-gradient evaluation via one L2 
regression (Sec.~\ref{sec:flowdpg}).

\begin{table*}[!t]
\centering
\small
\setlength{\tabcolsep}{2.2pt}
\caption{\small Full baseline sweep on AirPods assembly: per-stage,
overall, and average success over a single evaluation round of 25
cases (50 pod manipulations); per-stage rates count retries in the
denominator; Avg is the mean of the eight per-stage rates. Baselines
are trained offline only; Figure~\ref{fig:main_results} gives QAM the
same online phase as FlowDPG.}
\label{tab:main_airpods}
\begin{tabular}{l cccccccc | cc}
\toprule
\textbf{Method} &
\makecell{Grasp\\case} &
\makecell{Open\\case} &
\makecell{Grasp\\R pod} &
\makecell{Insert\\R pod} &
\makecell{Grasp\\L pod} &
\makecell{Insert\\L pod} &
\makecell{Close\\case} &
\makecell{Place\\case} &
\textbf{Overall} &
\textbf{Avg} \\
\midrule
BC (base)               & 100\% & 78.32\% & 72.44\% & 68.54\% & 78.56\% & 66.23\% & 92\%  & 96\%  & 64\% & 81.51\% \\
\midrule
\multicolumn{11}{l}{\emph{Value-conditioning}} \\
RA-BC                   & 100\% & 86.15\% & 80.52\% & 64.52\% & 82.49\% & 68.46\% & 96\%  & 96\%  & 76\% & 84.27\% \\
AWR                     & 100\% & 80.65\% & 78.43\% & 72.45\% & 78.77\% & 70.65\% & 100\% & 100\% & 76\% & 85.12\% \\
RECAP                   & 100\% & 89.29\% & 80.62\% & 69.69\% & 76.54\% & 64.52\% & 96\%  & 100\% & 72\% & 84.58\% \\
\midrule
\multicolumn{11}{l}{\emph{Auxiliary-module adaptation}} \\
DSRL                    & 100\% & 79.85\% & 75.47\% & 70.59\% & 78.13\% & 67.94\% & 92\%  & 92\%  & 68\% & 82.00\% \\
PLD                     & 100\% & 85.27\% & 80.13\% & 79.82\% & 80.43\% & 79.59\% & 96\%  & 96\%  & 76\% & 87.16\% \\
RLT                     & 100\% & 86.93\% & 82.41\% & 77.92\% & 84.14\% & 80.18\% & 96\%  & 100\% & 80\% & 88.45\% \\
\midrule
\multicolumn{11}{l}{\emph{Critic-gradient}} \\
QAM                     & 100\% & 88\%    & 83.33\% & 70.58\% & 73.52\% & 80\%  & 100\% & 100\%  & 80\%   & 86.93\%     \\
FlowDPG (offline)       & 100\% & 89.29\% & 90.63\% & 84.38\% & 96.43\% & 83.33\% & 100\% & 100\% & 88\% & 93.01\% \\
\textbf{FlowDPG (+online)} & \textbf{100\%} & \textbf{89.29\%} & \textbf{92.31\%} & \textbf{86.21\%} & \textbf{96.67\%} & \textbf{88.89\%} & \textbf{100\%} & \textbf{100\%} & \textbf{92\%} & \textbf{94.17\%} \\
\bottomrule
\end{tabular}
\end{table*}

\begin{figure*}[t]
\centering
\includegraphics[width=\linewidth]{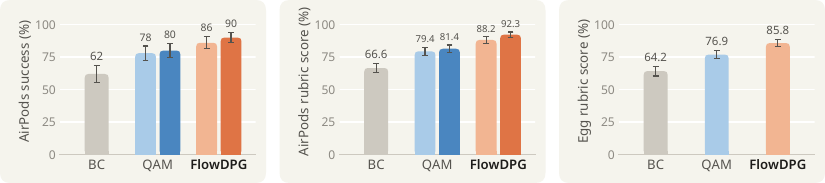}
\caption{\small Repeated evaluation against the BC base and 
the closest baseline QAM. Each bar is 50 rollouts (10 at each of 5 
checkpoints of one training run); error bars are the standard error 
over the five checkpoint means. Lighter shades: after the offline 
phase; darker shades: after online fine-tuning. The egg task is offline-only. Rubric: 0--3 
per stage as a percentage of the available points 
(Appendix~\ref{app:rubric}).}
\label{fig:main_results}
\end{figure*}

\paragraph{Repeated evaluation and a second task.} With the same 
online phase for both methods (Figure~\ref{fig:main_results}), FlowDPG 
leads QAM on AirPods by about ten points in both overall success and 
rubric score, and online fine-tuning helps FlowDPG more than QAM. Its 
90\% overall here is consistent with the 92\% of the single-round sweep 
(Table~\ref{tab:main_airpods}). On the egg task we compare only 
against BC and QAM, which ties for the best baseline on AirPods and is 
closest to FlowDPG in mechanism; the same ordering holds after the 
offline phase alone.
\subsection{Online RL Improves Robustness to Unseen Disturbances}
\label{sec:q2}

Online fine-tuning raises FlowDPG's overall success from 86\% to 90\%
in the repeated evaluation (Figure~\ref{fig:main_results}); in the
sweep, its per-stage gains all fall on the pod stages
(Table~\ref{tab:main_airpods}). To probe generalization beyond the
demonstration distribution, we apply three families of disturbances
after the policy reaches the relevant stage:
(i) \emph{stage undo}---returning a grasped case/pod to the tray or
inverting the open/close state of the case;
(ii) \emph{in-place adjustment}---dislodging a successfully
inserted pod; (iii) \emph{scene perturbation}---reshuffling tray
contents while the policy is reaching. Each disturbance is
evaluated on 50 trials.

\begin{table}[t]
\centering
\footnotesize
\caption{\small Recovery rates under disturbance families (i)--(iii); $\Delta$ in points.}
\label{tab:q2_disturbance}
\begin{tabular}{l cc c}
\toprule
\textbf{Disturbance} & \textbf{Offline} & \textbf{+Online} & $\Delta$ \\
\midrule
(i) Re-grasp (object replaced in tray)             & 78\% & 92\% & +14 \\
(i) Re-open/close (case state inverted)            & 82\% & 94\% & +12 \\
(ii) Re-insert (pod dislodged)                     & 86\% & 92\% & +6  \\
(iii) Adapt grasp (tray reshuffled mid-execution)  & 70\% & 88\% & +18 \\
\bottomrule
\end{tabular}
\end{table}

The offline policy handles undo disturbances reasonably 
(Table~\ref{tab:q2_disturbance}), since the post-undo observation 
resembles the start of the corresponding stage; its weak point is 
scene perturbation, a mid-execution transition absent from 
demonstrations. Online RL improves all four cases, most on scene 
perturbation, consistent with deployment-time failures exposing the 
policy to partial-success states that the critic-guided update learns 
to recover from.

\subsection{Diagnostics}
\label{sec:diagnostics}

A learned critic offers no guarantee outside the data distribution,
a limitation shared by DPG-style methods, and we do not explicitly
constrain the policy to the demonstration support, since exploring
beyond it is part of the purpose of RL. Online, an overvalued action
can produce an observable failure, such as a broken eggshell or an
improperly seated pod, which SARM registers and the critic learns
from, provided SARM remains accurate on newly visited states.
Table~\ref{tab:diagnostics} examines critic-gradient informativeness
and action-distribution changes on both tasks.

\emph{Critic-gradient informativeness:}
For each action chunk across 50 BC rollouts, we compute its
alignment with the critic gradient,
$\cos(\nabla_a Q(s,a),\, a-\mu_\pi(s))$,
where $\mu_\pi(s)$ is the mean of BC's sampled chunks at $s$,
and its increase in SARM progress.
We report their Spearman correlation ($\pm$ s.e.\ over rollouts).
BC was never optimized against $Q$, so this tests the gradient
on actions generated without critic guidance, against observed
progress rather than predicted value.
The positive correlations on both tasks suggest that the gradient
is locally informative within the BC visitation distribution,
but do not establish its reliability outside it.

\emph{Action-distribution changes:}
At 80 demonstration states, we sample $K{=}50$ action chunks
before and after RL using the same noise draws.
We measure their spread in per-dimension standardized action
units, their Wasserstein distance to demonstration actions at
nearby states, and their predicted values alongside those of
recorded demonstration actions
(with paired changes $\pm$ s.e.).
All values use the same final critic within each task,
isolating policy changes from changes in the scorer.
On both tasks, RL modestly narrows the action distribution
and slightly increases its distance to nearby demonstrations,
while raising predicted policy values above those of
demonstration actions.
Moving away from demonstrations is compatible with the purpose
of RL but does not by itself show improvement, and higher scores
from the critic being optimized against cannot separate genuine
gains from exploitation of critic errors, which small
distributional changes do not rule out. The task-success gains
above therefore remain the independent evidence of improvement.
\begin{table}[t]
\centering
\footnotesize
\setlength{\tabcolsep}{4pt}
\caption{\small Critic and action-distribution diagnostics on both tasks.}
\label{tab:diagnostics}
\begin{tabular}{l cc}
\toprule
 & \textbf{AirPods (Franka)} & \textbf{Eggs (YAM)} \\
\midrule
\multicolumn{3}{l}{\emph{Critic-gradient informativeness}} \\
Spearman $\rho$ & $0.36\pm0.05$ & $0.28\pm0.03$ \\
\midrule
\multicolumn{3}{l}{\emph{Action-distribution changes (before $\to$ after RL)}} \\
Standardized spread of samples & $1.61\to1.51$ \; ($-0.10\pm0.03$) & $0.87\to0.85$ \; ($-0.02\pm0.01$)\\
Wasserstein distance to nearby demos & $9.93\to10.02$ \; ($+0.09\pm0.04$) & $7.35\to7.48$ \; ($+0.13\pm0.05$) \\
$Q$ of policy actions & $32.79\to37.66$ \; ($+4.87\pm1.24$) & $53.84\to59.66$ \; ($+5.82\pm1.54$) \\
$Q$ of demonstration actions & $33.58$ & $54.24$ \\
\bottomrule
\end{tabular}
\end{table}

\subsection{Ablation: Consistency Regularizer and Adaptive Shift}
\label{sec:q3}

\begin{figure*}[t]
\centering
\includegraphics[width=\linewidth]{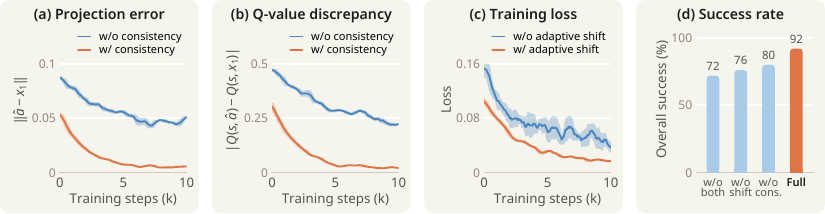}
\caption{\small Ablation on the consistency regularizer and adaptive
shift. Lines are a moving average, shading $\pm$1 SD. All variants
in (d) include the full online phase.}
\label{fig:q3_ablation}
\end{figure*}

We ablate each design choice and report both end-to-end success
and mechanism-level metrics (Figure~\ref{fig:q3_ablation}); all
variants are trained with the full online phase. Removing
the consistency regularizer lets the projection error
$\|\hata - x_1\|$ plateau an order of magnitude higher
(Fig.~\ref{fig:q3_ablation}(a)), amplifying the Q-value discrepancy
$|Q(s, \hata) - Q(s, x_1)|$ (Fig.~\ref{fig:q3_ablation}(b)). The critic gradient is then
evaluated farther from the data, where it is less reliable, and
Overall drops from 92\% to 80\%. Removing the
adaptive shift reduces the update to a raw $\alpha g$, whose
unbounded magnitude makes the training loss oscillate rather than
decrease smoothly (Fig.~\ref{fig:q3_ablation}(c)) and produces the largest
single-component drop (92\% to 76\%). The two components mitigate
the same root cause, a misdirected critic-gradient signal, through
complementary means, so removing both adds only a marginal further
degradation (72\%).

\subsection{Ablation: Reward Components}
\label{sec:q4}

\begin{figure}[t]
\centering
\includegraphics[width=\linewidth]{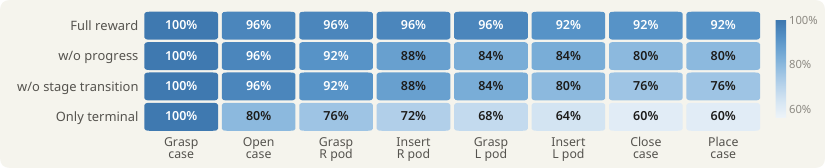}
\caption{\small Cumulative stage coverage on AirPods under reward 
ablation: the fraction of episodes that complete every stage up to 
and including each column.}
\label{fig:q4_stage_coverage}
\end{figure}

We compare the full composite reward in Eq.~\ref{eq:chunkreward}
against three variants; Figure~\ref{fig:q4_stage_coverage} reports
cumulative stage coverage with a distinct failure signature per
variant. Without the progress term, the reward is constant within
each stage except at transitions, so within-stage differences in $Q$
come only from discounting, $\nabla_a Q$ is weak, and coverage erodes
gradually across the sequence (80\%). Removing the stage-transition term keeps the
dense progress shaping but eliminates the discrete bonus at
sub-task completion, so $\nabla_a Q$ does not strongly favor
closing out a sub-task over hovering near its end (76\%). Removing both leaves only the
terminal bonus: two visually similar states, the gripper holding a
closed case before opening and again after the case is re-closed,
become indistinguishable under reward. The policy frequently
bypasses the entire pod-handling sequence to place the case
directly, dropping sharply at Open case ($96\%\!\to\!80\%$) and
ending at 60\%, below even the BC base. The progress and stage-transition terms are thus not redundant:
with such sparse shaping, the distilled critic-gradient signal
carries more noise than information.

\section{Conclusion and Limitations}
\label{sec:conclusion}

We presented \textbf{FlowDPG}, a DDPG-style method for flow 
matching policies that distills the critic gradient into the 
velocity field via a single-step projection and L2 regression, 
avoiding backpropagation through the denoising ODE and recovering 
vanilla DPG under three explicit approximations. On two long-horizon 
real-world tasks, it attains 92\% end-to-end success on dual-arm 
AirPods assembly and an 85.8\% rubric score on 14-stage egg cooking.

\paragraph{Limitations.} FlowDPG relies on a learned critic, whose 
gradient is most reliable near the data. The one-step estimate and the 
adaptive shift reduce, but do not eliminate, the effect of critic 
errors, and online interaction corrects them only as far as the reward 
model scores newly visited states accurately. Our reward assumes tasks 
that decompose into stages with a dense progress signal, and the BC 
anchor that keeps actions feasible may also limit how far the policy 
can improve. Finally, our evaluation covers two long-horizon tasks on 
two robot platforms; extending FlowDPG to larger VLA policies and a 
broader range of tasks is left for future work.

\clearpage
\acknowledgments{We thank Ananye Agarwal and Kevin Gmelin for helpful discussions. We are
grateful to the Skild~AI manipulation and hardware teams for their support, and to all the
contractors who collected and annotated the data used in this work.}

\bibliography{example}  %
\clearpage
\appendix
\section{Derivation of the Vanilla DPG Gradient on a Flow Matching Policy}
\label{app:vanilla_dpg}

We provide the full derivation of Eq.~\ref{eq:vanilla_dpg} for 
completeness. The deterministic actor is the ODE terminus 
$x_1 = x_0 + \int_0^1 \vth(x_t, t, s)\, dt$, with $x_0 \sim \NN(0, I)$ 
independent of $\theta$. The DDPG actor loss is 
$\LL_{\text{DPG}} = -Q(s, x_1)$, so by the chain rule
\begin{equation}
    \nabla_\theta \LL_{\text{DPG}} 
    = -\nabla_a Q(s, x_1) \cdot \nabla_\theta x_1
    = -g^\top \cdot \nabla_\theta x_1,
    \label{eq:app_dpg_chain}
\end{equation}
where $g := \nabla_a Q(s, a)|_{a = x_1} \in \RR^{|a|}$ is the critic 
gradient at the action terminus, treated throughout as a column 
vector. It remains to derive $\nabla_\theta x_1$ for the ODE actor.

\paragraph{Forward sensitivity equation.} Define the sensitivity matrix 
$M(t) := \partial x_t / \partial \theta \in \RR^{|a| \times |\theta|}$, 
with initial condition $M(0) = 0$ since $x_0$ is parameter-free. 
Differentiating the ODE $\frac{d x_t}{d t} = \vth(x_t, t, s)$ with 
respect to $\theta$ and applying the chain rule on the right-hand side,
\begin{equation}
    \frac{d M(t)}{d t} 
    = \nabla_x \vth(x_t, t, s) \cdot M(t) + \nabla_\theta \vth(x_t, t, s).
    \label{eq:app_sensitivity_ode}
\end{equation}
This is a linear non-homogeneous ODE in $M(t)$. By variation of 
parameters, its solution at $t{=}1$ is
\begin{equation}
    M(1) = \int_0^1 \Phi(1, t) \cdot \nabla_\theta \vth(x_t, t, s)\, dt,
    \label{eq:app_sensitivity_sol}
\end{equation}
where $\Phi(\sigma, t)$ is the state-transition matrix associated with 
the homogeneous part of Eq.~\ref{eq:app_sensitivity_ode}, defined as 
the solution to
\begin{equation}
    \frac{\partial \Phi(\sigma, t)}{\partial \sigma} 
    = \nabla_x \vth(x_\sigma, \sigma, s) \cdot \Phi(\sigma, t), 
    \qquad \Phi(t, t) = I.
    \label{eq:app_phi_ode}
\end{equation}
$\Phi(1, t)$ describes how a perturbation of $x_t$ propagates to a 
perturbation of $x_1$ — i.e., the Jacobian $\partial x_1 / \partial x_t$.

\paragraph{Final form.} Substituting Eq.~\ref{eq:app_sensitivity_sol} 
into Eq.~\ref{eq:app_dpg_chain} gives the vanilla DPG gradient quoted 
in the main text:
\begin{equation}
    \nabla_\theta \LL_{\text{DPG}} 
    = -\int_0^1 g^\top \Phi(1, t) \cdot \nabla_\theta \vth(x_t, t, s)\, dt.
    \label{eq:app_vanilla_dpg_final}
\end{equation}

\paragraph{Discrete-time form.} Under Euler discretization 
$x_{t_{i+1}} = x_{t_i} + \Delta t \cdot \vth(x_{t_i}, t_i, s)$ with 
$T$ steps and $\Delta t = 1/T$, the chain-rule recursion gives
\begin{equation}
    \nabla_\theta x_1 
    = \Delta t \cdot \sum_{i=0}^{T-1} \bigg[\prod_{j=i+1}^{T-1} J_j\bigg] \cdot \nabla_\theta \vth(x_{t_i}, t_i, s),
    \quad J_j := I + \Delta t \cdot \nabla_x \vth(x_{t_j}, t_j, s),
\end{equation}
with the convention $\prod_{j=T}^{T-1}(\cdot) = I$, so that the term 
$i = T{-}1$ contributes only $\nabla_\theta \vth(x_{t_{T-1}}, \cdot)$. 
The matrix product $\prod_{j=i+1}^{T-1} J_j$ is the discrete analogue 
of $\Phi(1, t_i)$. In practice this recursion is what an autograd 
implementation would compute when backpropagating through the ODE 
solver. The product structure is precisely the source of the gradient 
explosion/vanishing pathology: small deviations in the spectral radius 
of $J_j$ from unity compound multiplicatively across the $T$ steps, 
making the update numerically fragile when $\vth$ is not yet 
well-trained.

\paragraph{Adjoint form.} Equivalently, define the adjoint state 
$\lambda(t) := \Phi(1, t)^\top g \in \RR^{|a|}$, which propagates the 
terminal gradient $g$ backward in time. It satisfies the linear 
backward ODE
\begin{equation}
    \frac{d \lambda(t)}{d t} 
    = -\nabla_x \vth(x_t, t, s)^\top \cdot \lambda(t), 
    \qquad \lambda(1) = g,
\end{equation}
integrated from $t = 1$ to $t = 0$. The DPG gradient can then be 
written as
\begin{equation}
    \nabla_\theta \LL_{\text{DPG}} 
    = -\int_0^1 \lambda(t)^\top \cdot \nabla_\theta \vth(x_t, t, s)\, dt.
\end{equation}
This is the form that adjoint methods solve directly, computing 
$\lambda(t)$ by backward integration and accumulating $\nabla_\theta$ 
contributions at each $t$. FlowDPG bypasses both the forward 
sensitivity matrix $\Phi$ and the adjoint state $\lambda$ entirely, 
instead distilling a single critic gradient evaluated at the 
in-distribution projected clean action $\hata$.

\section{Offline-to-Online Training Algorithm}
\label{app:algorithm}

Algorithm~\ref{alg:main} summarizes the full offline-to-online procedure of Sec.~\ref{sec:pipeline}.

\begin{algorithm}[h]
\small
\caption{FlowDPG: Offline-to-Online Training}
\label{alg:main}
\begin{algorithmic}[1]
\Require Offline buffer $\DD_{\text{off}}$; online buffer $\DD_{\text{on}}$ (initially empty); guidance scale $\alpha$; consistency weight $\mu_{\text{cons}}$; max distillation weight $\lambda_{\max}$; warmup $N_{\text{warmup}}$; EMA rate $\rho$; expectile $\tau$.
\State Pretrain SARM on a small stage-annotated subset of the data
\State Initialize backbone, action head $\vth$, value head $(\Vpsi, Q_{\phi_1}, Q_{\phi_2})$; EMA targets $Q_{\bar\phi_i} \gets Q_{\phi_i}$
\Statex \hspace*{-\algorithmicindent}\makebox[\dimexpr\linewidth+\algorithmicindent][c]{\midrulefill\quad\textbf{Offline phase}\quad\midrulefill}
\For{$k = 1, \ldots, N_{\text{off}}$}
    \State Sample mini-batch from $\DD_{\text{off}}$; compute $r_t$ from the reward predictor
    \State Sample $x_0 \sim \NN(0, I)$, $t \sim \mathrm{Uniform}(0, 1)$; form $x_t, u_t$
    \State Compute $v_t = \vth(x_t, t, s)$ and $\hata$ via Eq.~\ref{eq:project}
    \State $\LL_{\text{cons}} \gets \|\hata - x_1\|^2$
    \State $g \gets \nabla_{\hata} \min_{i} Q_{\phi_i}(s, \hata)$ with stop-grad on critic
    \State $\Delta \gets \alpha \cdot \|u_t\| / (\|g\| + \varepsilon) \cdot g$
    \State $\astar \gets \hata + \Delta$; $\utstar \gets \astar - x_0$
    \State $\lambda \gets \lambda_{\max} \cdot \min(k / N_{\text{warmup}}, 1)$
    \State Update $\theta$ by minimizing $\LL_{\text{FlowDPG}} + \mu_{\text{cons}} \LL_{\text{cons}}$ via Eqs.~\ref{eq:flowdpgloss},~\ref{eq:cons}
    \State Update $\psi, \phi_1, \phi_2$ via Eqs.~\ref{eq:vloss}--\ref{eq:qloss}
    \State EMA targets: $\bar\phi_i \gets \rho \bar\phi_i + (1-\rho)\phi_i$
\EndFor
\Statex \hspace*{-\algorithmicindent}\makebox[\dimexpr\linewidth+\algorithmicindent][c]{\midrulefill\quad\textbf{Online phase}\quad\midrulefill}
\State Deploy $\pi_\theta$ for autonomous rollout collection
\For{$k = 1, \ldots, N_{\text{on}}$}
    \State Asynchronously append rollouts to $\DD_{\text{on}}$ with rewards from the reward predictor
    \State Sample mini-batch from $\DD_{\text{off}} \cup \DD_{\text{on}}$ at $\sim 1{:}1$ ratio; BC anchor on demos and successful rollouts only
    \State Update $\theta, \psi, \phi_1, \phi_2$ as in the offline phase, with backbone and SARM frozen
    \If{$k \bmod N_{\text{sync}} = 0$}
        \State Synchronize $\pi_\theta$ to deployment
    \EndIf
\EndFor
\State \textbf{return} $\vth$
\end{algorithmic}
\end{algorithm}

\newpage

\section{SARM Signals and Value Estimates Along a Rollout}
\label{app:sarm_visualization}

\begin{figure*}[ht]
\centering
\includegraphics[width=\linewidth,
                 trim=438 142 279 139, clip]{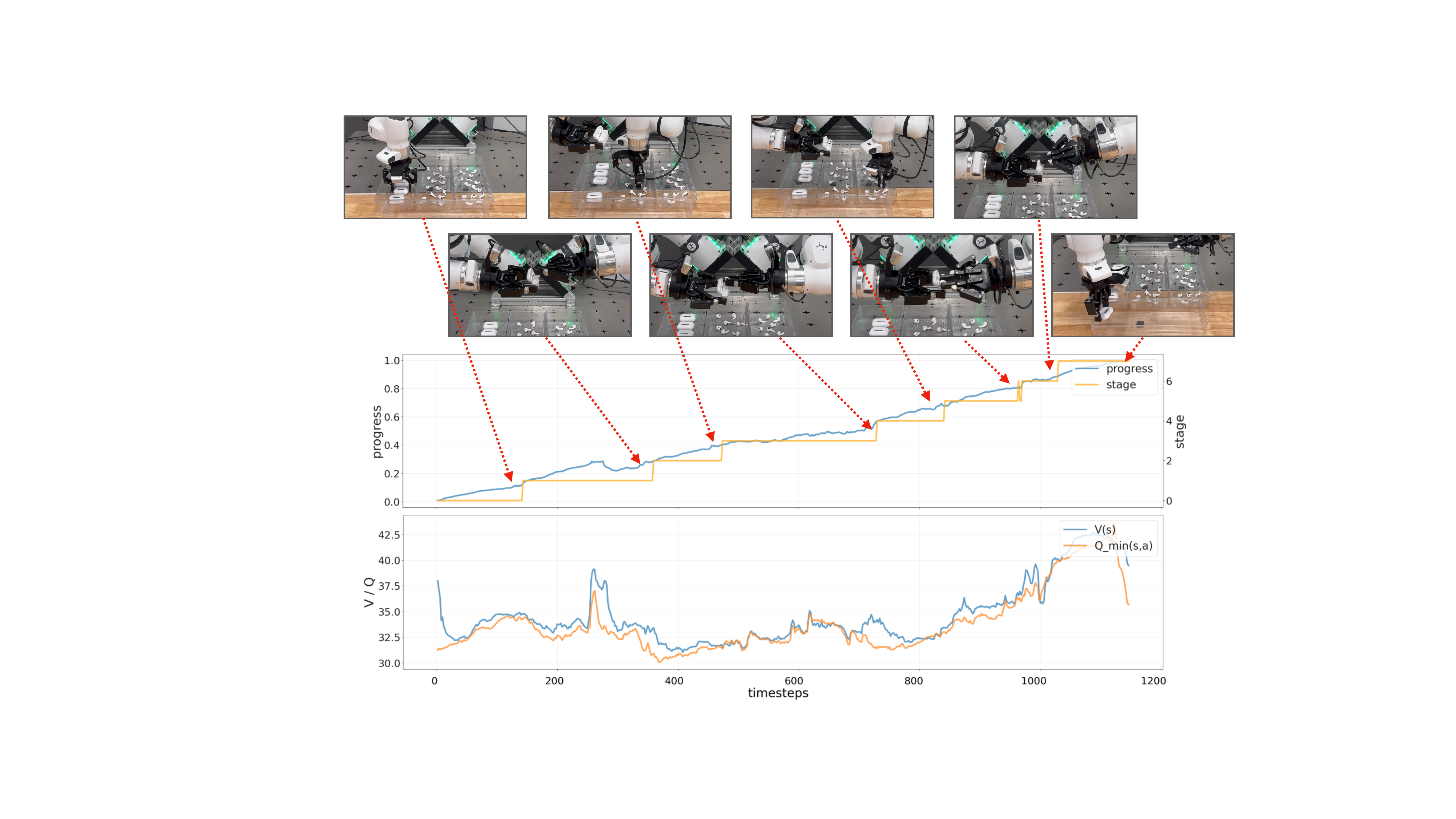}
\caption{SARM signals and value estimates along a successful AirPods 
assembly rollout. \textbf{Top:} keyframes at SARM stage transitions. 
\textbf{Middle:} predicted within-stage progress $\phi(o_t)$ (blue) 
and stage label $z(o_t)$ (yellow step). \textbf{Bottom:} learned 
value $\Vpsi(s_t)$ (blue) and twin-critic minimum 
$\min_i Q_{\phi_i}(s_t, a_t)$ (orange) over the same rollout.}
\label{fig:sarm_value}
\end{figure*}

Figure~\ref{fig:sarm_value} visualizes how the SARM signals and the 
learned value/Q-estimates evolve along a successful real-world 
execution. The progress curve $\phi(o_t)$ provides a meaningful 
per-step signal even within a single stage, supplying the dense 
shaping that FlowDPG distills via the critic, and stage-label jumps 
align with the keyframes in the top strip, illustrating when the 
stage-transition bonus in Eq.~\ref{eq:chunkreward} fires. The value 
and Q-curves track each other closely throughout the rollout 
($\Vpsi \approx \min_i Q_{\phi_i}$), indicating that the value and 
critic heads are mutually consistent. The value trends upward over the 
rollout, mostly in the later stages, with transient dips at 
contact-rich stages that coincide with short recoveries from minor 
failures.

\clearpage
\section{Evaluation Rubrics}
\label{app:rubric}

Each stage is scored 0--3; a stage that is never reached scores 0. 
The criteria below were fixed before scoring. The rubric score is the 
total as a percentage of the available points: 24 for AirPods 
assembly (8 stages) and 42 for scrambled-egg cooking (14 stages). In 
the egg task, \emph{pick up egg} and \emph{crack egg} occur twice, once 
per egg, giving 14 scored stages from the 12 stage types in 
Table~\ref{tab:rubric_egg}.

\newcolumntype{Y}{>{\raggedright\arraybackslash}X}
\begin{table}[h]
\centering
\scriptsize
\setlength{\tabcolsep}{3pt}
\renewcommand{\arraystretch}{1.12}
\caption{\small Rubric for AirPods assembly (Franka), 8 stages, 24 points.}
\label{tab:rubric_airpods}
\begin{tabularx}{\linewidth}{>{\raggedright\arraybackslash}p{1.55cm} Y Y Y Y}
\toprule
\textbf{Stage} & \textbf{3 -- Successful} & \textbf{2 -- Minor error} & \textbf{1 -- Major error} & \textbf{0 -- Failed} \\
\midrule
Grasp case & Secure grasp, correct orientation, no slip & Grasp with adjustment or delay & Case lifted but shifted or dropped once & Fails to grasp case \\
Open case & Lid fully opened in one clean motion & Opens with adjustment or hesitation & Opens only after repeated attempts, or case displaced & Lid not opened \\
Grasp right pod & Secure grasp, correct approach direction & Grasp with adjustment or delay & Wrong approach direction, pod dislodged then re-grasped & Fails to grasp pod \\
Insert right pod & Seated on the first attempt, fully in & Seated after one retry & Seated after two or more retries, or insertion dislodges the other pod & Pod not seated \\
Grasp left pod & Secure grasp, correct approach direction & Grasp with adjustment or delay & Wrong approach direction, pod dislodged then re-grasped & Fails to grasp pod \\
Insert left pod & Seated on the first attempt, fully in & Seated after one retry & Seated after two or more retries, or insertion dislodges the right pod & Pod not seated \\
Close case & Lid closed cleanly with both pods seated & Closes after minor adjustment & Closes with a pod unseated, or needs repositioning & Case not closed \\
Place case & Placed upright within the target region & Placed with minor misalignment & Dropped or placed outside the region & Case not placed \\
\bottomrule
\end{tabularx}
\end{table}

\begin{table}[h]
\centering
\scriptsize
\setlength{\tabcolsep}{3pt}
\renewcommand{\arraystretch}{1.12}
\caption{\small Rubric for scrambled-egg cooking (YAM), 14 stages, 42 points.}
\label{tab:rubric_egg}
\begin{tabularx}{\linewidth}{>{\raggedright\arraybackslash}p{1.55cm} Y Y Y Y}
\toprule
\textbf{Stage} & \textbf{3 -- Successful} & \textbf{2 -- Minor error} & \textbf{1 -- Major error} & \textbf{0 -- Failed} \\
\midrule
Pick up egg & Secure grasp, no slip or damage & Grasp with adjustment or delay & Lifts egg but drops or damages it & Fails to grasp egg \\
Crack egg & Clean crack, proper shell separation & Minor shell fragments or awkward motion & Incorrect or messy crack & Fails to crack egg \\
Put salt & Correct amount applied accurately & Minor spillage or hesitation & Incorrect amount or placement & Salt not applied, or bottle dropped \\
Pick up fork & Secure and correct grasp & Grasp with adjustment & Unstable or incorrect grasp & Fails to pick up fork, or drops it \\
Whisk egg & Thorough whisking with correct motion & Partial or inefficient whisking & Minimal or incorrect motion & No whisking \\
Place fork back & Placed neatly in correct location & Minor misalignment & Dropped or poorly placed & Fork not placed back \\
Put eggs into pan & Clean transfer into pan & Minor spillage or hesitation & Significant spillage & Eggs not transferred \\
Pick up spatula & Secure and correct grasp & Grasp with adjustment & Unstable or incorrect grasp & Fails to pick up spatula \\
Stir eggs & Proper stirring across pan & Incomplete or uneven stirring & Minimal or incorrect stirring & No stirring \\
Transfer eggs & Clean and accurate transfer & Minor spillage or inefficiency & Partial or messy transfer & Eggs not transferred \\
Place spatula back & Placed correctly and safely & Minor misplacement & Dropped or unsafe placement & Spatula not placed back \\
Serve eggs & Served correctly and fully & Minor error or delay & Incorrect or incomplete serving & Eggs not served \\
\bottomrule
\end{tabularx}
\end{table}

\end{document}